\documentclass{article}

\usepackage{microtype}
\usepackage{amssymb,amsmath}
\usepackage{graphicx}
\usepackage{subfigure}
\usepackage{booktabs} 

\usepackage{lipsum}
\usepackage{multicol}

\usepackage{hyperref}

\usepackage[accepted]{icml2020}

\icmltitlerunning{Selective Dyna-Style Planning Under Limited Model Capacity}

\begin{document}

\twocolumn[
\icmltitle{Selective Dyna-Style Planning Under Limited Model Capacity}

\begin{icmlauthorlist}
\icmlauthor{Zaheer Abbas}{alb}
\icmlauthor{Samuel Sokota}{alb}
\icmlauthor{Erin J. Talvitie}{hm}
\icmlauthor{Martha White}{alb}
\end{icmlauthorlist}

\icmlaffiliation{alb}{The University of Alberta and the Alberta Machine Intelligence Institute (Amii)}
\icmlaffiliation{hm}{Harvey Mudd College}

\icmlcorrespondingauthor{Zaheer Abbas}{mzaheer@ualberta.ca}
\icmlcorrespondingauthor{Samuel Sokota}{sokota@ualberta.ca}

\icmlkeywords{Machine Learning, ICML}

\vskip 0.3in
]

\printAffiliationsAndNotice{} 

\begin{abstract}
In model-based reinforcement learning, planning with an imperfect model of the environment has the potential to harm learning progress.
But even when a model is imperfect, it may still contain information that is useful for planning.
In this paper, we investigate the idea of using an imperfect model \textit{selectively}. 
The agent should plan in parts of the state space where the model would be helpful but refrain from using the model where it would be harmful.
An effective selective planning mechanism requires estimating predictive uncertainty, which arises out of aleatoric uncertainty, parameter uncertainty, and model inadequacy, among other sources.
Prior work has focused on \textit{parameter uncertainty} for selective planning.
In this work, we emphasize the importance of \textit{model inadequacy}.
We show that heteroscedastic regression can signal predictive uncertainty arising from model inadequacy that is complementary to that which is detected by methods designed for parameter uncertainty, indicating that considering both parameter uncertainty and model inadequacy may be a more promising direction for effective selective planning than either in isolation.
\end{abstract}

\section{Introduction}

Reinforcement learning is a computational approach to learning via interaction. 
An algorithmic agent is tasked with determining a policy that yields a large cumulative reward.
Generally, the framework under which this agent learns its policy falls into one of two groups: model-free reinforcement learning or model-based reinforcement learning.
In model-free reinforcement learning, the agent acts in ignorance of any explicit understanding of the dynamics of the environment, relying solely on its state to make decisions.
In contrast, in model-based reinforcement learning, the agent possesses a \textit{model} of how its actions affect the future.
The agent uses this model to reason about the implications of its decisions and \textit{plan} its behavior. 

The model-based approach to reinforcement learning offers significant advantages in two regimes. 
The first is domains in which acquiring experience is expensive.
Model-based methods can leverage planning to do policy improvement without requiring further samples from the environment.
This is important both in the traditional Markov decision process setting, where sample efficiency is often an important performance metric,
and also in a more general pursuit of artificial intelligence, where an agent may need to quickly adapt to new goals.
Second is the regime in which capacity for function approximation is limited and the optimal value function and policy cannot be represented.
In such cases, agents that plan at decision-time can construct temporary local value estimates whose accuracy exceed the limits imposed by capacity restriction \cite{silver2008sample}. 
These agents are thereby able to achieve policies superior to those of similarly limited model-free agents.

Far from being special cases, sample-sensitive, limited-capacity settings are typical of difficult problems in reinforcement learning.
It is therefore not surprising that many of the most prominent success stories of reinforcement learning are model-based.
In the Arcade Learning Environment (ALE) \cite{bellemare2013arcade}, algorithms that distribute training across many copies of an exact model of the environment have been shown to massively outperform algorithms limited to a single instance of the environment \cite{r2d2}.
And in Chess, Shogi, Go, and Poker, superhuman performance can be reached by means of decision-time planning on exact transition models \cite{silver2018general,deepstack,libratus}.

However, the premise of these successes is not the same as that of the classical reinforcement learning problem. 
Rather than being asked to learn a model from interactions with a black box environment, these agents are provided an exact model of the dynamics of the environment.
While the latter is in itself an important problem setting, the former is more central to the pursuit of broadly intelligent agents.

Unfortunately, learning a useful model from interactions has proven difficult. 
While there are some examples of success in domains with smooth dynamics \cite{pilco,planet}, learning an accurate model in more complex environments, such as the ALE, remains difficult.
In a pedagogical survey of the ALE, \citeauthor{marlos} (\citeyear{marlos}) state ``So far, there has been no clear demonstration of successful planning with a learned model in the ALE.''
Until recently \cite{muzero}, basic non-parametric models that replay observed experiences \cite{per} remained convincingly superior to state-of-the-art parametric models \cite{van2019use}.

Some of the difficulty of increasing performance with a learned model arises from the multifold nature of the issue.
Learning a useful model is itself a challenge.
But even once a model is learned, it is not clear when and how an imperfect model is best used.
The use of an imperfect model can be catastrophic to progress if the model is incorrectly trusted by the agent \cite{talvitie2017self, jafferjee2020hallucinating}.

This work concerns itself with how to effectively use an imperfect model.
We discuss planning methods that only use the model where it makes accurate predictions. 
Such techniques should allow the agent to plan in regions of the state space where the model is helpful, but refrain from using the model when it would be damaging. 
We refer to this idea as \textit{selective planning}. 

There are two interrelated problems involved in selective planning: determining when the model is and is not accurate, and devising a planning algorithm which uses that information to plan selectively.

We formulate the first problem as that of predictive uncertainty estimation and consider three sources of predictive uncertainty---aleatoric uncertainty, parameter uncertainty, and model inadequacy---emphasizing the relevance of model inadequacy for selective planning under limited-capacity.
We demonstrate that the learned input-dependent variance \cite{nix1994estimating} can reveal the presence of predictive uncertainty that is not captured by standard tools for quantifying parameter uncertainty.

We address the second problem by empirically investigating selective planning in the context of model-based value expansion (MVE), a planning algorithm that uses a learned model to construct multi-step TD targets \cite{feinberg2018model}.
The results show that MVE can fail when the model is subject to capacity constraints. 
In contrast, we find that selective MVE, an instance of selective planning that weights the multi-step TD targets according to the uncertainty in the model's predictions, can perform sample-efficient learning even with an imperfect model that otherwise leads to planning failures.

\section{Background}

This section provides background on model-based reinforcement learning, sources of predictive uncertainty, and previous work exploiting estimates of parameter uncertainty for model-based reinforcement learning. 

\subsection{Model-Based Reinforcement Learning} 
Reinforcement learning problems are typically formulated as a finite Markov decision processes (MDPs).
An MDP is defined by a tuple $(\mathcal{S}, \mathcal{A}, r, p)$, where $\mathcal{S}$ is the set of states, $\mathcal{A}$ is the set of actions, $r \colon \mathcal{S} \times \mathcal{A} \times \mathcal{S} \to \mathbb{R}$ is the reward function, and $p \colon (s_t, a_t, s_{t+1}) \mapsto P(S_{t+1}{=}s_{t+1} \mid S_t{=}s_t, A_t{=}a_t)$ is the dynamics function.
At each time-step $t$, the environment is in some \textit{state} $s_t \in \mathcal{S}$, the agent executes an \textit{action} $a_t \in \mathcal{A}$, and the environment transitions to state $s_{t+1} \in \mathcal{S}$ and emits a reward $r_{t+1} = r(s_t, a_t, s_{t+1})$.
The agent acts according to a \textit{policy} $\pi: \mathcal{S} \to \Delta(\mathcal{A})$, which maps states to probabilities of selecting each possible action (we use $\Delta(X)$ to denote the simplex on $X$).
The agent may maintain this policy explicitly or derive it from a value function $v \colon \mathcal{S} \to \mathbb{R}$ or an action-value function $q \colon \mathcal{S} \times \mathcal{A} \to \mathbb{R}$, which predict an expected discounted cumulative reward, given the state and state-action pair, respectively.
The agent's goal is to use its experience to learn a policy that maximizes expected discounted cumulative reward.

In model-based reinforcement learning, the agent leverages a model capturing some aspects of the dynamics of the environment.
This work regards the problem setting in which the agent must learn this model from its experience, rather than being endowed with it a priori.
In particular, the experiments in this work investigate learning models of the form $m \colon \mathcal{S} \times \mathcal{A} \to \mathbb{R}^{k} \supset \mathcal{S}$ (we assume $\mathcal{S}$ is embedded in the standard $k$-dimensional Euclidean space for some positive integer $k$).
Such models deterministically predict the expected next state from the current state and action.
While there is nothing that constrains their predictions to the state space and they are unable to express non-deterministic transitions, models of this form can still offer useful information.

Dyna \cite{sutton1991dyna} is an approach to model-based reinforcement learning that combines learning from real experience and experience simulated from a learned model.
The characterizing feature of Dyna-style planning is that updates made to the value function and policy do not distinguish between real and simulated experience.
In this work, we investigate the idea of selective Dyna-style planning.
An effective selective planning mechanism should focus on states and actions for which the model makes accurate predictions.

\subsection{Sources of Predictive Uncertainty}

Predictive uncertainty, the aggregate uncertainty in a prediction, arises from {\em aleatoric uncertainty} (due to randomness intrinsic to the environment), {\em parameter uncertainty} (due to uncertainty about which set of parameters generated the data), and model inadequacy, among other sources.
This section discusses the situations in which these sources of uncertainty appear in the context of model-based reinforcement learning.

\noindent\textbf{Aleatoric Uncertainty:} 
In reinforcement learning, aleatoric uncertainty comes from the dynamics function $p$.
If the dynamics function induces non-deterministic transitions (those which occur with probability greater than zero but less than one), the agent cannot be certain what transition will occur.
Aleatoric uncertainty is irreducible in the sense that it cannot be resolved by collecting more samples or increasing the complexity of the model.

\noindent\textbf{Parameter Uncertainty:} 
Parameter uncertainty is the uncertainty over the values of the parameters, given a parametric hypothesis class and the available data.
In model-based reinforcement learning, this is a result of the finite dataset used to train the model.
This dataset will not contain a transition for every state and action.
And for stochastic transitions, even if a state-action pair has been sampled multiple times, it is unlikely to have been sampled frequently enough to accurately reflect the underlying distribution.
These insufficiencies cause uncertainty in the sense that is unclear which parameter values are correct.
Unlike aleatoric uncertainty, parameter uncertainty can be reduced (and in the limit, eliminated \citep{de1937prevision}) by gathering more data.

Bayesian inference is a common approach to estimating parameter uncertainty.
However, analytically computing the posterior over parameters is intractable for large neural networks.
A significant body of research on Bayesian neural networks is concerned with approximating this posterior \citep{mackay1992bayesian, hinton1993keeping, barber1998ensemble, graves2011practical, gal2016dropout, gal2017concrete, li2017dropout}.

The statistical bootstrap is an alternative line of research for estimating parameter uncertainty \citep{efron1982jackknife}. 
These methods train an \textit{ensemble} of neural networks, possibly on independent \textit{bootstrap samples} of the original training samples, and use the empirical parameter distribution of the ensemble to estimate parameter uncertainty \citep{lakshminarayanan2017simple, osband2016deep, pearce2018uncertainty, osband2018randomized}.
Ensemble-based methods can be interpreted as Bayesian approximations only under restricted settings \citep{fushiki2005nonparametric, fushiki2005bootstrap, osband2018randomized}, but share the goal of quantifying uncertainty due to insufficient data.

\noindent\textbf{Model Inadequacy:}
Model inadequacy refers to the model's hypothesis class being unable to express the underlying function generating the data.
In reinforcement learning, it is typical that the true dynamics function is not an element of the model's hypothesis class, both because this functional form is unknown and because the dynamics functions can be very complex.
Thus, even in the limit of infinite data, the parameter values that most accurately fit the dataset may not accurately predict transitions.
Error due to model inadequacy can only be resolved by increasing the capacity of the model.

\begin{figure}
   \vspace{0.7cm}
  \includegraphics[width=0.5\textwidth]{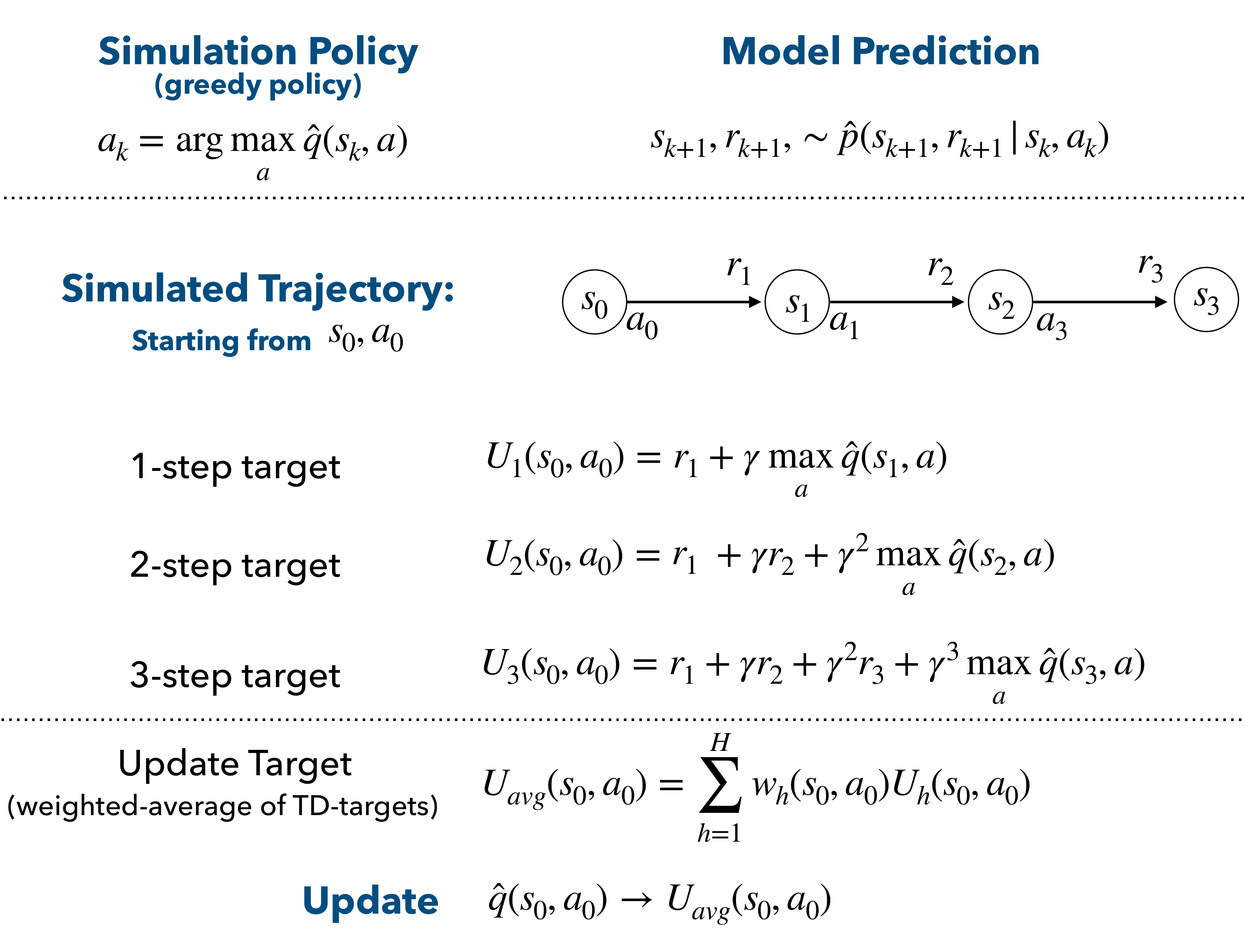}
  \caption{A depiction of MVE. A 3-step trajectory is simulated using an approximate model $\hat{p}$ from a state-action pair $(s_0, a_0)$; the simulated trajectory is used to construct multi-step TD targets; the TD-targets are combined using a weighted average; the action-value function $\hat{q}$ is updated toward the weighted average $U_{avg}$.}
   \label{fig:pictorial_mve}
\end{figure}

\subsection{Selective Planning with Ensembles}
Selective planning requires that the agent possess a mechanism for deciding when to use the model.
While ensembling is not the only existing approach (e.g. \citeauthor{xiao2019learning}) to designing this mechanism, it is the most prominent \cite{abbeel, kalweit}.
A particularly relevant selective planning method that uses ensembling is stochastic ensemble value expansion (STEVE) \citep{buckman2018sample}.
STEVE estimates parameter uncertainty by augmenting model-based value expansion (MVE) \cite{feinberg2018model}, an extension of DQN \cite{mnih2015human} in which model-simulated experience is used to evaluate the greedy policy, as is shown in Figure \ref{fig:pictorial_mve}.
STEVE uses the degree of agreement among an ensemble of neural networks to approximate the trustworthiness of the model for a particular rollout length.
Rollout lengths with low variance are given more weight in the update and rollout lengths with high variance are given less.

\begin{figure*}
\begin{center}
\includegraphics[width=\linewidth]{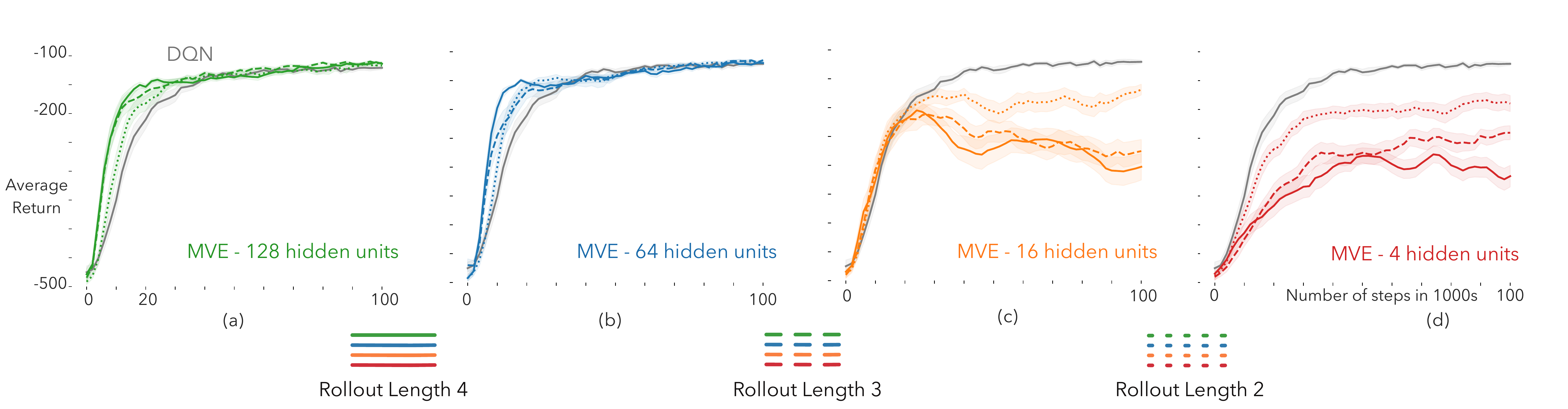}
      \vspace{-0.7cm}
  \caption{The performance of MVE in Acrobot, for varying model capacity, averaged over 30 runs with shaded region corresponding to standard error.
  When the model is given sufficient capacity to express the dynamics, MVE can increase performance, as is shown in the left two plots. However, when the model is lacking capacity, MVE can catastrophically damage learning, as is shown in the right two plots.}
  \label{fig:mve-fail}
  \end{center}
\end{figure*}

\section{Limited Capacity Can Harm Planning}\label{sec:mve_harm}

A main hypothesis of this work is that neglecting model inadequacy during planning can cause catastrophic failure.
To establish this idea, we begin by presenting an experiment examining the relationship between model capacity and performance in Acrobot \cite{sutton1996generalization}, a classic environment loosely based on a gymnast swinging on a highbar.
We ran MVE with four different network architectures for the value function:
one hidden layer with 64 hidden units, one hidden layer with 128 hidden units, two hidden layers with 64 hidden units each, and two hidden layers with 128 hidden units each.
For each network architecture, we determined the best setting for the step size, the batch size, and the replay memory size by sweeping over possible parameter configurations, as detailed in Appendix \ref{app:sweep}.

The results for the value function with one hidden layer with 128 hidden units, shown in Figure \ref{fig:mve-fail}, suggest that the relationship between capacity and performance is as anticipated.
When given sufficient capacity to learn a good model, MVE has the potential to improve the sample efficiency of DQN.
However, as capacity is decreased and the model becomes unable to accurately reflect the underlying dynamics, MVE harms learning progress.
The results for the other value function architectures, which can be found in Appendix \ref{app:baseline}, tell similar stories.

\section{Estimating Predictive Uncertainty Due to Limited Capacity Using Heteroscedasticity} \label{sec:error_limited_capacity}

While neural networks of reasonable sizes are perfectly capable of expressing the Acrobot dynamics function, this may not be the case in highly complex domains.
To defend against this possibility, it is desirable to have a mechanism for detecting predictive uncertainty due to model inadequacy.

We hypothesize that methods effective at detecting input-dependent noise should also be able to estimate predictive uncertainty due to model inadequacy.
The intuition behind this hypothesis is that a complex function can equally validly be considered a simple function with complex disturbances.
For example, $f \colon x \mapsto x + \sin(x)$ can be viewed as a linear function with input-dependent, deterministic disturbances.
Thus, methods designed for heteroscedastic regression may also quantify predictive uncertainty due to model inadequacy.
In contrast, parameter uncertainty methods may overlook predictive uncertainty due to model inadequacy and instead simply agree on the best parameter values within the hypothesis class in the limit of data.

\subsection{Heteroscedastic Regression} \label{sec:learned_variance_gaussian}
Neural networks are typically trained to output a point estimate as a function of the input. 
When trained with mean-squared error, the probabilistic interpretation is that the point estimate corresponds to the mean of a Gaussian distribution with fixed input-independent variance $\sigma^2$: $p(y | \mathbf{x}) = \mathcal{N}(f_\mu(\mathbf{x}), \sigma^{2})$; 
maximizing the likelihood in this case leads to least-squares regression. 

An alternative is to assume that the variance is also input-dependent: $p(y | \mathbf{x}) = \mathcal{N}(f_\mu(\mathbf{x}), f_{\sigma^2}(\mathbf{x}))$, where $f_\mu(\mathbf{x})$ is the predicted mean and $f_{\sigma^2}(\mathbf{x})$ is the predicted variance.  Under this assumption, maximizing the likelihood leads to the following loss function \citep{nix1994estimating}:
\begin{equation}
\label{eq:heteroscadastic}
L_i(\theta) = \frac{(y_i - f_\mu(\mathbf{x}_i))^2}{2f_{\sigma^2}(\mathbf{x}_i)} + \frac{1}{2}\text{log}f_{\sigma^2}(\mathbf{x}_i).
\end{equation}
The learned variance $f_{\sigma^2}(\mathbf{x})$ can be predictive of stochasticity. 
The network can incur less penalty in high-noise regions of the input space by predicting high variance. 
We hypothesize that this learned variance should also be predictive of the errors in the context of limited capacity---the network can maintain a small loss by allowing the variance to be larger in regions where it lacks the capacity to make accurate predictions.

\subsection{Experimental Setup} \label{sec:regressionExample}
To examine this hypothesis, we contrast a subset of parameter uncertainty methods with heteroscedastic regression on a simple regression problem.

\begin{figure}
\hspace{-2.0cm}
  \begin{center}\includegraphics[width=0.5\textwidth]{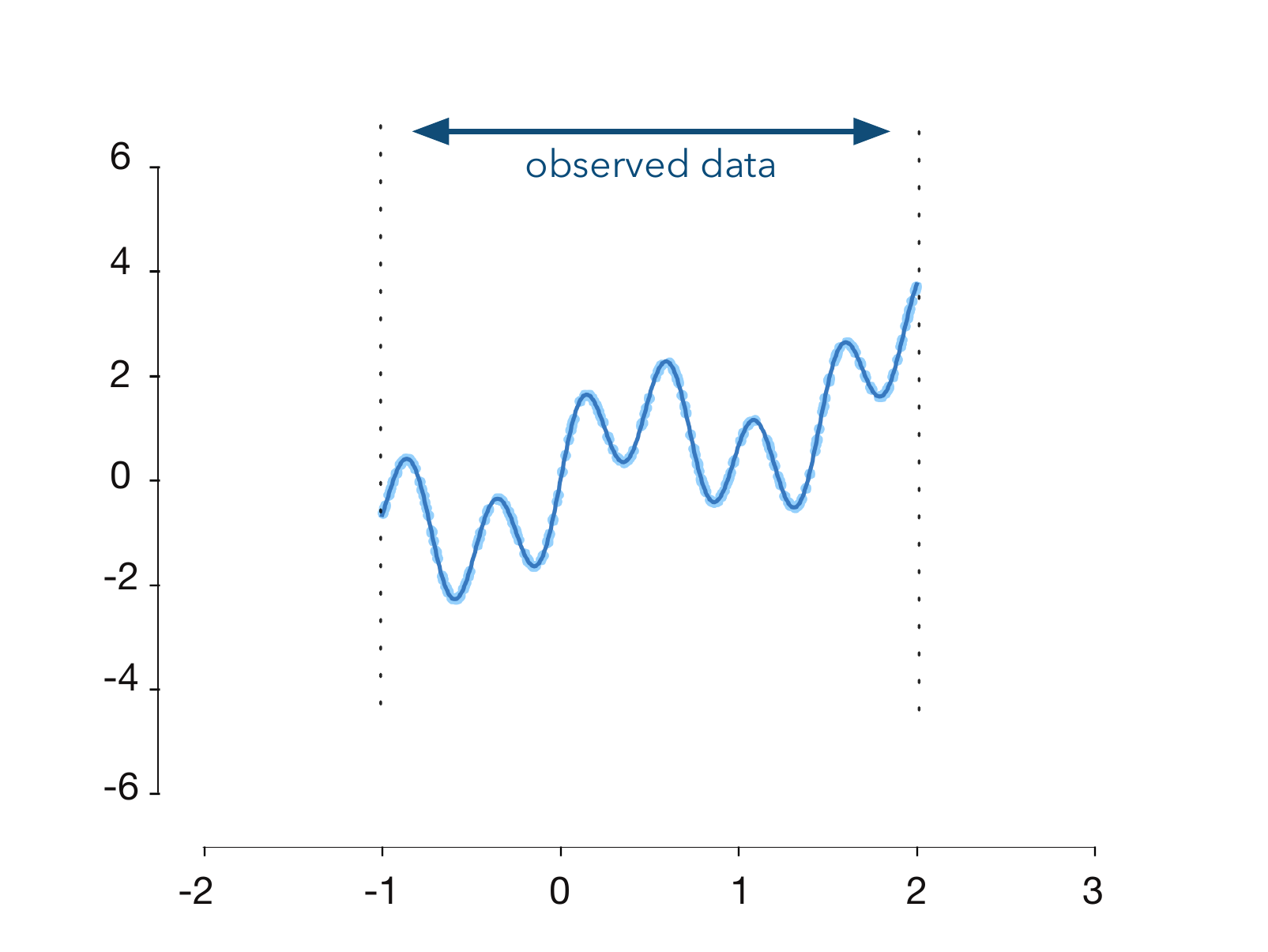}\end{center}
      \vspace{-0.7cm}
  \caption{The target function $y = x + \sin(4 x) + \sin(13 x)$ shown for the training interval (-1.0, 2.0). The blue points are 300 training samples drawn uniform randomly from the training interval.
  }
  \label{fig:regression_example_fig}
\end{figure}

We constructed a dataset of 5,000 training examples using the function $y = x + \sin(\alpha x) + \sin(\beta x)$, where $\alpha=4$, $\beta=13$, and $x$ was drawn uniformly from the interval $(-1.0, 2.0)$ (see Figure \ref{fig:regression_example_fig}). 
The inputs $x$ are drawn from a uniform distribution over the interval $(-1.0, 2.0)$.

We applied neural networks to this regression problem and varied the effective capacity of the model by reducing the number of layers and the number of hidden units. 
In particular, we used neural networks with three degrees of complexity:  3 hidden layers with 64 hidden units each (referred to as large network), a single hidden layer with 2048 hidden units (medium-size network), and a single hidden layer with 64 hidden units (small network). 
Experimental details are described in Appendix \ref{app:regression}.

We compared heteroscedastic regression with the following approaches for estimating parameter uncertainty.

\begin{figure*}
\begin{center}
\includegraphics[width=\linewidth]{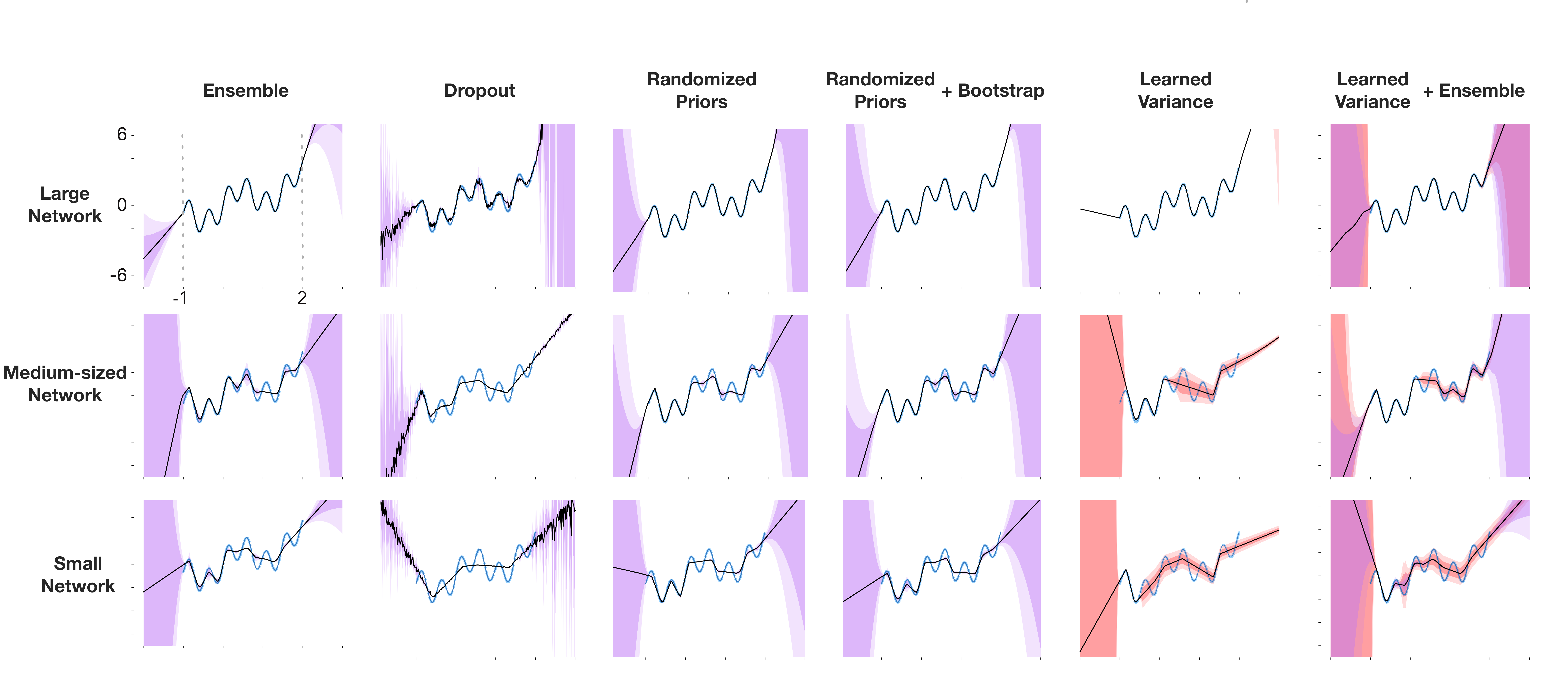}
\caption{An evaluation of uncertainty methods on a simple regression problem when the model is subject to capacity limitations.
The ground truth function is in blue in all plots. 
Each row presents the mean predictions and uncertainty estimates for a particularly-sized neural network after training for 300 epochs; each column presents the results for a particular uncertainty method.
Uncertainty estimates are represented by shaded intervals; the estimated predictive uncertainty arising from parameter uncertainty is in purple; the learned variance is in red; darker purple/red intervals show mean $\pm$ 1 standard-deviation and lighter intervals show mean $\pm$ 2 standard-deviation. Learning rate is $0.001$ for all methods; the results for other configurations of the learning rate (consistent with the results presented here) can be found in Appendix \ref{app:regression}.
}\label{fig:regression_results}
\end{center}
\end{figure*}

\subsubsection{Monte Carlo Dropout}
\citeauthor{gal2016dropout} (\citeyear{gal2016dropout}) proposed to use \textit{dropout} \cite{srivastava2014dropout} for obtaining uncertainty estimates from neural networks. Dropout is a regularization method which prevents overfitting by randomly dropping units during training.  Monte Carlo dropout estimates uncertainty by computing the variance of the predictions obtained by $M$ stochastic forward passes through the network. This technique can be interpreted through the lens of Bayesian inference; that is, the dropout distribution approximates the Bayesian posterior \cite{gal2016uncertainty}.

\subsubsection{Ensemble of Neural Networks} \label{sec:ensemble_nn}
Ensembling independently trains $K$ randomly initialized neural networks. 
The variance in the predictions of individual networks is used to estimate predictive uncertainty arising from parameter uncertainty \cite{lakshminarayanan2017simple, osband2016deep, pearce2018uncertainty}. Intuitively, the individual networks in an ensemble should make similar predictions in the regions of the input space where sufficient samples have been observed, while making dissimilar predictions elsewhere.

\subsubsection{Randomized Prior Functions} \label{sec:rpf}
Randomized prior functions (RPF) \cite{osband2018randomized} are an extension of ensembling. 
Each network in the ensemble is coupled with a random but fixed \textit{prior} function---a randomly initialized neural network whose weights remain unchanged during training. 
The prediction of an individual ensemble member is the sum of its trainable network and its prior function.
For Gaussian linear models, this approach is equivalent to exact Bayesian inference \cite{osband2018randomized}.

\subsubsection{Randomized Prior Functions With Bootstrapping} \label{sec:rpf_b}
Bootstrapping can be combined with both randomized prior functions \cite{osband2018randomized} and ensembling \cite{osband2016deep}.
We focus on the former as it has been noted to provide better uncertainty estimates \cite{osband2018randomized}.

\subsection{Experimental Results}

The results of applying each of the above methods to the regression problem are shown in Figure \ref{fig:regression_results} for a single configuration of the learning rate. We found the results to be consistent across learning rate configurations (see Appendix \ref{app:regression} for additional results).

With a sufficiently powerful network (large network), the ensemble learns to accurately predict the target function, and the predictive variance of the ensemble (purple) appropriately assesses the predictive uncertainty---the ensemble variance is large outside the training distribution. 
We observe the same effect for other parameter uncertainty methods. 

As the capacity is reduced (small and medium-sized networks), all methods fail to fit the target function accurately over the entire input space.
But whereas learned variance reliably reflects the errors within the training distribution, the parameter uncertainty methods fail to do so. 
These results support the idea that parameter uncertainty is insufficient for selective planning in the face of limited capacity. 
Instead, they suggest that a combination of a parameter uncertainty and model inadequacy would yield a more robust error detection mechanism than either individually, as is indicated by the rightmost column, which combines learned variance with ensembling.

\section{Selective Planning} \label{sec:sec_selective_mve}

In this section, we investigate the utility of learned variance in the context of model-based reinforcement learning; 
in particular, we ask whether learned variance can be used to plan selectively with a low-capacity model that otherwise leads to the planning failures observed in Section \ref{sec:mve_harm}.

Toward this end, we describe a technique using learned variance to do selective model-based value expansion.
Given a maximum rollout length $H$, consider the weighted-average of h-step targets (see Figure \ref{fig:pictorial_mve}):
\begin{equation*}
\label{eq:h-step-target-generic}
U_{avg}(s_0, a_0) = \sum_{h=1}^{H} w_h(s_0, a_0) U_h(s_0, a_0).
\end{equation*}
We would like the weight on an h-step target to be inversely related to the cumulative uncertainty $\sigma_{1:h}^2(s_0, a_0) = \sum_{i=1}^h \sigma_i^2(s_0, a_0)$ over the h-step trajectory.
Given the cumulative uncertainty of the targets, we determine the weight of an individual target by computing the softmax
\begin{equation}
\label{eq:h-step-target}
w_h(s_0, a_0)  = \frac{\text{exp} (-\sigma_{1:h}^2(s_0, a_0)/\tau)}{\sum_{i=1}^{H} \text{exp} (-\sigma_{1:i}^2(s_0, a_0)/\tau)}
\end{equation}
where $\tau$ is a hyper-parameter which regulates the weighting's sensitivity to the predicted uncertainty. 

To handle the multidimensional output space, we assume independence across different dimensions of the state (i.e., an isotropic Gaussian assumption) and learn a separate variance for each dimension.
To acquire a scalar value, we sum the variance values across the dimensions of the state space (i.e., we use the trace of the diagonal covariance matrix).
Further implementation details, including the range of values for the parameter sweep, and the configuration of the rest of the hyper-parameters are included in Appendix \ref{app:selective}.

\begin{figure*}
\begin{center}
\includegraphics[width=1.02\linewidth]{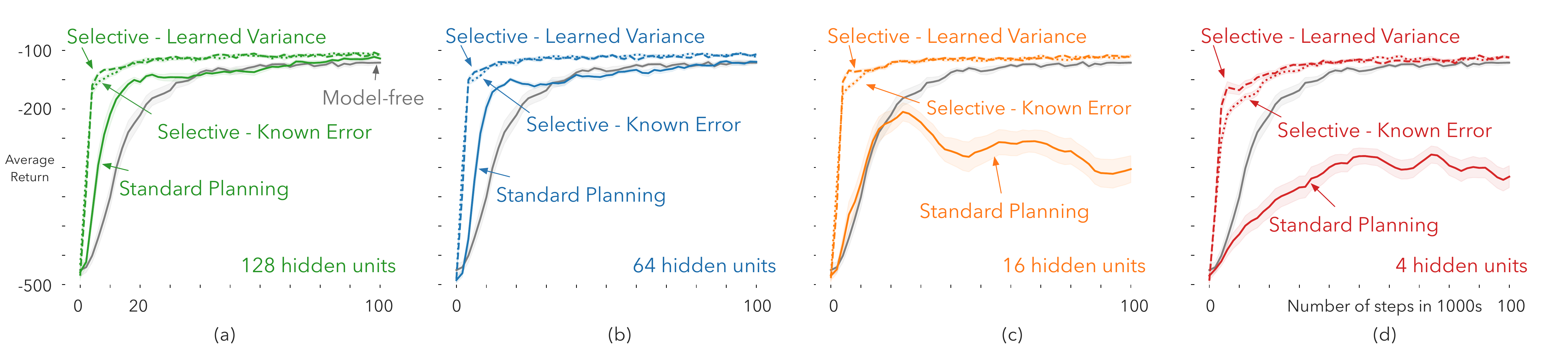}
  \vspace{-0.7cm}
  \caption{Results of selective MVE ($\tau=0.1$). The learning curves are averaged over 30 runs; the shaded regions show the standard error. Selective MVE with models of 4 hidden units (d) and 16 hidden units (c) not only matches the asymptotic performance of DQN, but also achieves better sample-efficiency than the DQN baseline. Selective MVE improves the sample-efficiency even in the case of larger models consisting of 64 hidden units (b) and 128 hidden units (a). }
  \end{center}
  \label{fig:acrobot_selectivemve_results}
\end{figure*}
\subsection{Selective MVE Avoids Catastrophic Failure}

In this section, we apply the planning algorithm described above, which we call selective MVE, to Acrobot.
We compare selective MVE using learned variance to standard MVE, DQN with no model-based updates, and selective MVE using the true squared error (given by an oracle) to weight its rollouts. 
The results, presented in Figure \ref{fig:acrobot_selectivemve_results}, show that selective MVE under capacity constraints (models with 4 and 16 hidden units) not only matches the asymptotic performance of DQN, effectively avoiding planning failures, but is also more sample-efficient than the DQN baseline. 

Interestingly, selective MVE improves sample-efficiency even in the case of larger models consisting of 64 and 128 hidden units.
This may indicate that selective planning allows the agent to make effective use of the model early in training when many of its predictions are unreliable but some are accurate.

\begin{figure}[h]
    \hspace{-7mm}
  \includegraphics[width=0.55\textwidth]{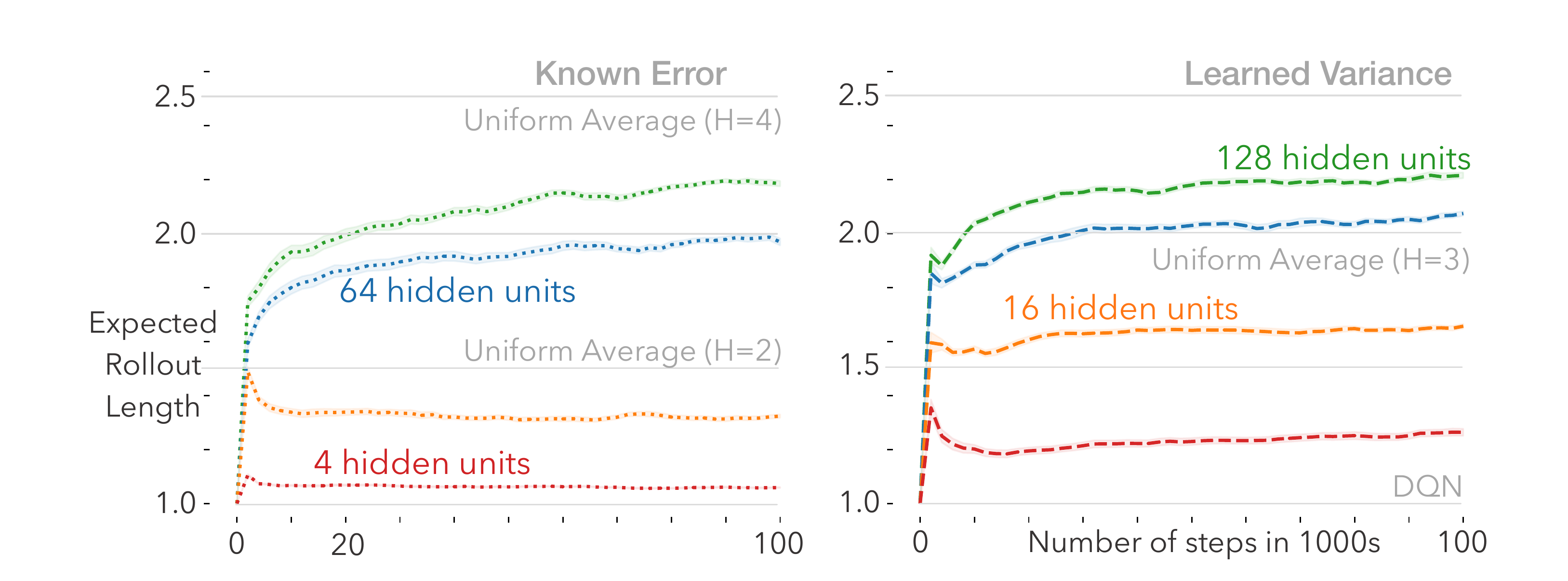}
    \vspace{-0.2cm}
  \caption{The plots contrast the expected rollout length of selective MVE ($\tau=0.1$) for the known error (left) and the learned variance (right). Each reported curve is the average of 30 runs; the shaded regions show the standard error.}
  \label{fig:acrobot_selectivemve_exp_rollout}
\end{figure}

To get a better of sense of the robustness of selective MVE to model errors, we compute the expected rollout length of each of the four model sizes for both the learned variance and the true error, as is shown in Figure \ref{fig:acrobot_selectivemve_exp_rollout}. 
(The expected rollout length of an update is the weighted average over the rollout lengths used in equation \ref{eq:h-step-target}.)
There is a clear ordering in the expected rollout length of the four models; 
$h$--step targets consisting of longer trajectories are given relatively more weight when the model is larger (and as a result, more accurate). 
Selective MVE with the smallest model of 4 hidden units does not use the model as much as its variants with bigger models, but the limited use is still sufficient to improve the sample efficiency of DQN, while preventing model inaccuracies from hurting control performance. 

\begin{figure}
\vspace{-0.7cm}
\begin{center}
  \includegraphics[width=0.5\textwidth]{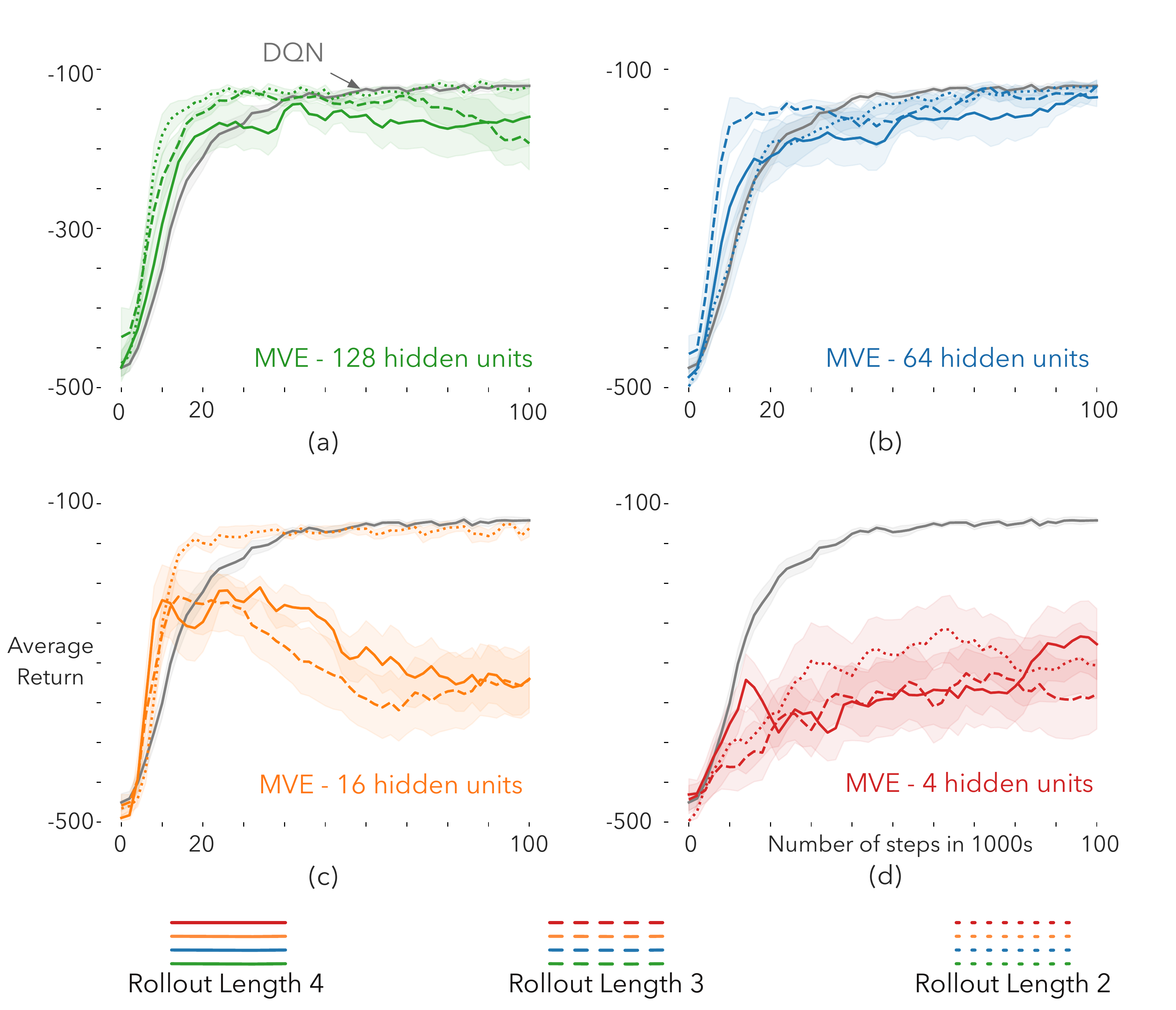}
  \vspace{-0.55cm}
  \caption{Performance of MVE when the model is learned using the heteroscedastic regression's loss function. The learning curves are averaged over 10 runs; the shaded regions show the standard error.}
  \label{fig:loss-plot}
  \end{center}
  \vspace{-0.5cm}
\end{figure}

\subsubsection{Improved Performance Cannot be Attributed to the Loss Function}
To verify that the gains in performance are not due to the change in loss function (selective MVE uses a heteroscedastic regression loss function, whereas MVE uses MSE---a homoscedastic loss function) we evaluated MVE with the same loss function as that of selective MVE. 
The results, presented in Figure \ref{fig:loss-plot}, suggest that simply changing the loss function does not lead to an accurate model, and that the model still needs to be used selectively.

\subsubsection{Improved Performance Cannot be Replicated With Ensembling}
\begin{figure}
\begin{center}
  \includegraphics[width=0.5\textwidth]{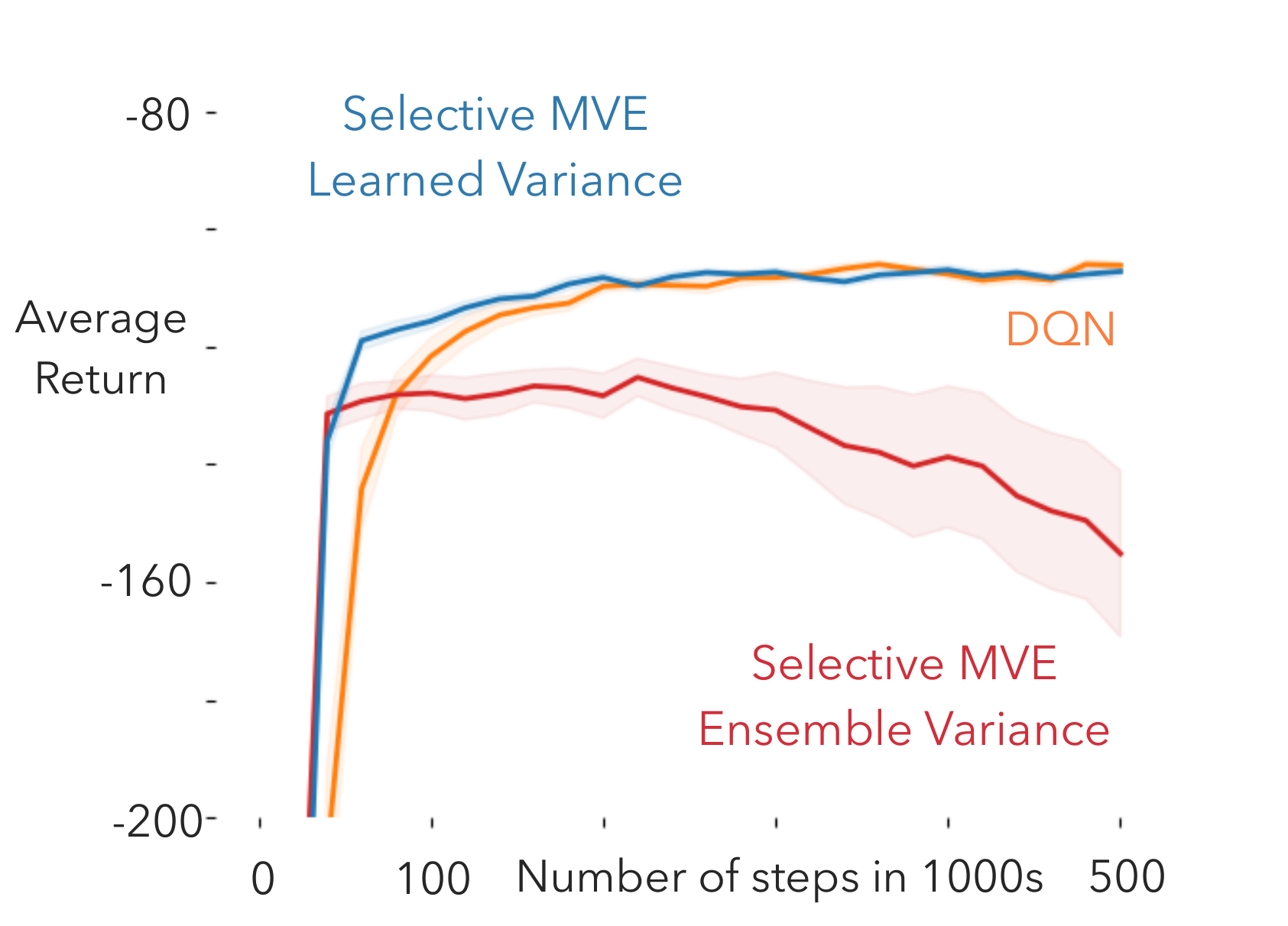}
    \vspace{-0.1cm}
  \caption{Results of selective MVE ($\tau=0.1$) over 500 thousand steps. The learning curves are averaged over 30 runs; the shaded regions show the standard error. Selective MVE with ensemble variance initially offers performance competitive with that of learned variance, but ultimately collapses.}
  \label{fig:ensemble-fail}
  \end{center}
\end{figure}

To verify that the same performance gains could not be achieved from ensembling, we applied a variant of selective MVE using ensemble variance, rather than learned variance.
Selective MVE with ensemble variance resembles STEVE, except that it uses variance over state predictions instead of over value predictions and softmax instead of inverse weightings.
While selective MVE with ensemble variance performs comparably to selective MVE with learned variance early in training, the performance of the planner using ensemble variance consistently collapses later in training, presumably as a result of the ensemble converging to similar, incorrect parameter values. 
Results for the architecture with 4 hidden units are shown in Figure \ref{fig:ensemble-fail}.

\section{Complementary Nature of Parameter Uncertainty and Model Inadequacy}
\begin{figure}[t]
\begin{center}
  \includegraphics[width=0.5\textwidth]{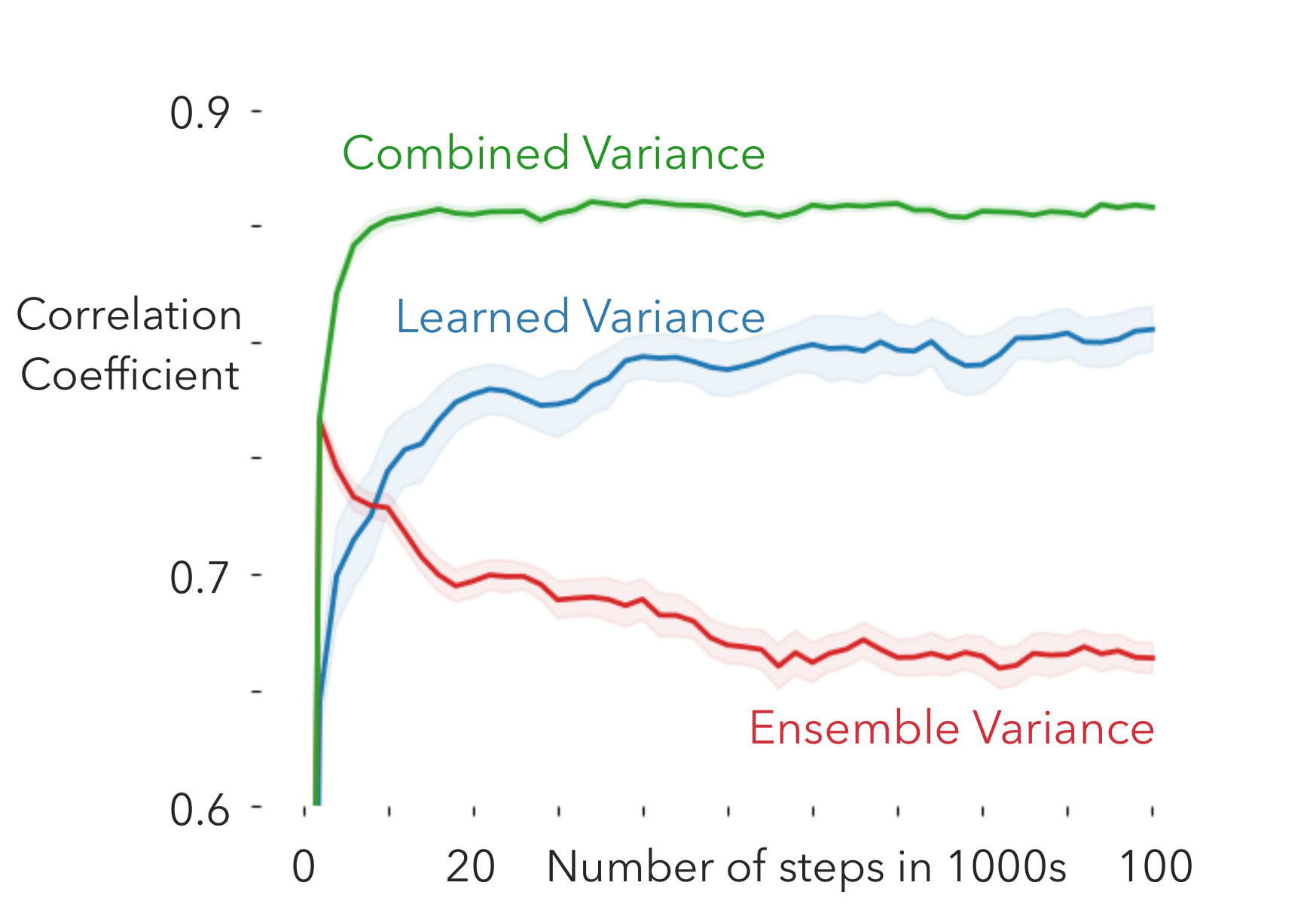}
  \vspace{0.22cm}
  \caption{Correlation of the true squared error with the learned variance, the ensemble variance, and the combined variance over the course of the agent's training. Each reported curve is the average of 30 runs; the shaded regions show the standard error.}
  \label{fig:correlation_plot}
  \end{center}
\end{figure}

In Section \ref{sec:error_limited_capacity}, we argued that predictive uncertainty arising from parameter uncertainty is by itself is insufficient for selective planning under capacity constraints, and needs to be used in combination with predictive uncertainty arising from model inadequacy. 
In Section \ref{sec:sec_selective_mve}, we used learned variance to estimate predictive uncertainty arising from model inadequacy, and found that it can ensure robust planning under capacity limitations. 
In this section, we emphasize the complementary nature of parameter uncertainty and model inadequacy.

We extend our Acrobot example and train an ensemble of neural networks with heteroscedastic loss functions. 
We use an ensemble of 20 single hidden layer networks with 4 hidden units, and
use the mean value of the ensemble to make a prediction.
To compute the variance, we consider the ensemble as a uniform mixture over Gaussians, along each dimension.
We compute the variance of the mixture model along each dimension and sum the variances as we did with heteroscedastic regression to get a scalar value. 

To evaluate the efficacy of the combined variance relative to learned variance and ensembles variance, we sample a batch of transitions from the replay buffer at every time-step and compute the correlation of each variance with the true mean squared error.
The results, shown in Figure \ref{fig:correlation_plot}, suggest that in the context of limited capacity: 1) Ensemble variance becomes a less useful indicator of error as training progresses, presumably because the ensemble tends to converge to similar predictions. 
2) Learned variance becomes more predictive of error as it learns from more data. 
3) Combined variance is more strongly correlated with true error than both learned variance and ensemble variance over the entire course of training.
While existing work \cite{sergey} has investigated combining heteroscedastic regression with ensembling in the context of non-deterministic domains, our results suggest that doing so has positive benefits under capacity limitations even in the absence of stochasticity.

\section{Conclusion}
In this work, we investigated the idea of selective planning: the agent should plan only in parts of the state space where the model is accurate.
We highlighted the importance of \textit{model inadequacy} for selective planning under limited model capacity.
Our experiments suggest that heteroscedastic regression, under an isotropic Gaussian assumption, can reveal the presence of error due to model inadequacy, whereas methods for quantifying parameter uncertainty do not do so reliably.
In the context of model-based reinforcement learning, we show that incorporating learned variance into planning can outperform the equivalent model-free method, even when using the model non-selectively would lead to catastrophic failure.
Lastly, we offer evidence that ensembling and heteroscedastic regression have complementary strengths, suggesting that their combination is a more robust selective planning mechanism than either in isolation.

\section{Acknowledgements}
This material is based upon work supported in part by the National Science Foundation under Grant No. IIS1939827, the Alberta Machine Intelligence Institute, and the Canadian Institute For Advanced Research.

\bibliography{paper}
\bibliographystyle{icml2020}
\clearpage
\newpage
\appendix

\newpage
\section{Parameter Sweep Strategy}\label{app:sweep}
To determine the best-performing hyper-parameter setting, we evaluate each configuration using 10 independent runs initialized with a different random seed, leading to as many learning curves.
We averaged the learning curves and summed the second half of the averaged learning curves to obtain a single number representing the performance of the particular configuration. 
If the best-performing parameter setting falls on the boundary of the range of tested values for any hyper-parameter, we widened the range until this was not true. 
We evaluated the best-performing configuration using 30 additional runs, each initialized with a different random seed.
We report the average learning curve, along with its standard error.
In all experiments, the resolution of the reported curves is 2,000 steps: we log the average returns of the most recent 20 episodes every 2,000 steps.

\section{Baseline Algorithms} \label{app:baseline}
\subsection{Deep Q-Networks (DQN)}
We estimate DQN's action-value function using a fully connected neural network with ReLU activations. 
We repeat all experiments with four different fully connected neural network architectures: 1 hidden layer with 64 hidden units, 1 hidden layer with 128 hidden units, 2 hidden layers with 64 hidden units each, and 2 hidden layers with 128 hidden units each. For each network architecture, we determine the best setting for the step-size, the batch size, and the replay memory size by sweeping over possible parameter configurations. For DQN baseline, the range of values for the parameter sweep, and the configuration of the rest of the hyper-parameters are presented in Table \ref{tab:acrobot_dqn_configuration}.

\subsection{Model-Based Value Expansion (MVE)}
We implement MVE by extending the DQN algorithm with the model-based policy evaluation technique described in Figure 1 of the main paper.
We instantiate MVE with a deterministic model learned using the squared error loss.
We assume the reward signal to be known; that is, we only learn the dynamics function. 
We study the effect of model capacity by progressively reducing the size of the neural network used for model learning. 
In particular, we use four variants of a single hidden layer neural network, which vary only in the number of hidden units: 128 hidden units, 64 hidden units, 16 hidden units, and 4 hidden units. 
In all cases, we learn the model online using the experience gathered in the replay buffer: at every time-step, alongside the MVE value function update, we separately sample a batch of transitions to update the model. 

Once we have identified the best hyper-parameter configuration for the DQN baselines which vary in their value function architecture, we keep the same hyper-parameter configuration for their MVE extensions, and only sweep over the model learning rate for each of the four model architectures. The range of values for the parameter sweep, and the configuration of the rest of the hyper-parameters are presented in Table \ref{tab:acrobot_mve_configuration}.

\begin{figure*}[h]
  \includegraphics[width=0.99\textwidth]{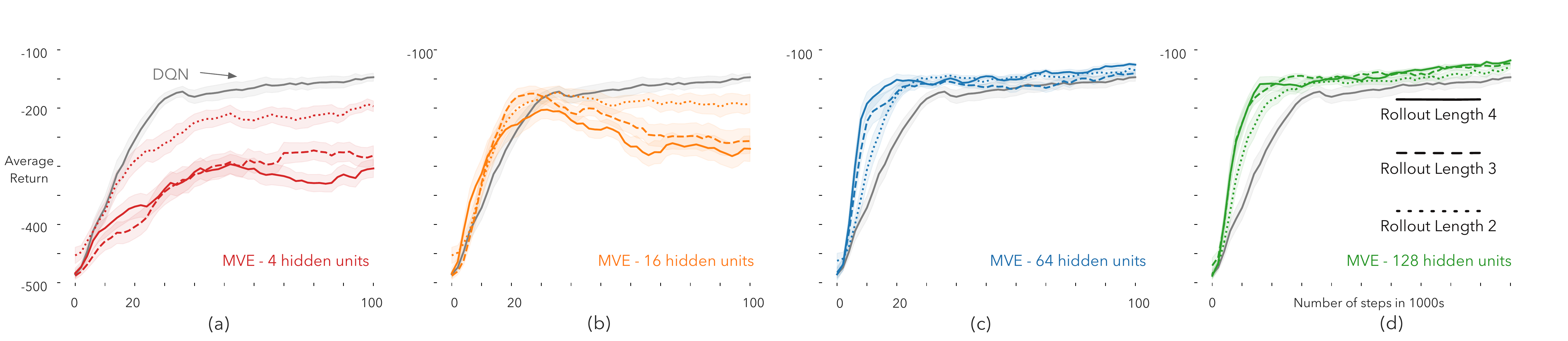}
  \caption{The effect of model capacity on MVE's performance with the value function network consisting of a single hidden layer and 64 hidden units.}
  \label{fig:vf_64h_effect_k}
\end{figure*}
\begin{figure*}[h]
  \includegraphics[width=0.99\textwidth]{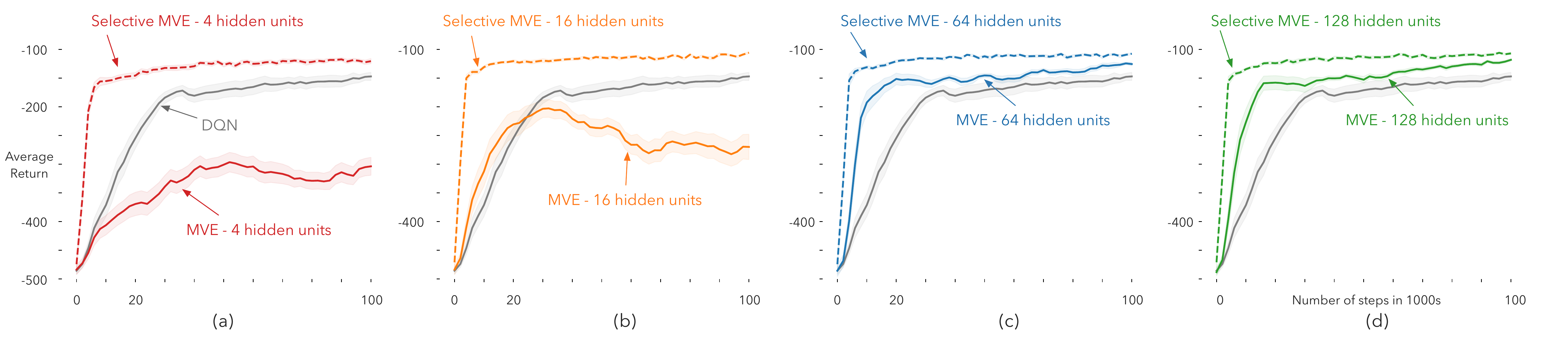}
  \caption{Results of selective MVE ($\tau=0.1$) for the value function network consisting of a single hidden layer and 64 hidden units.}
  \label{fig:vf_64h_holistic}
 \end{figure*}

\section{Selective Model-based Value Expansion: Additional Details and Results}\label{app:selective}

\noindent\textbf{Learned Variance Implementation Details: } We modify the base neural networks for the dynamics function to output the diagonal covariances alongside the mean next-state vector.  We enforce the positivity constraint on the covariances by passing the corresponding output through the \textit{softplus} function $\log(1+ \exp(\cdot))$; and, for numerical stability, we also add a small constant value of $10^{-6}$ to the predicted covariances \citep{lakshminarayanan2017simple}. The model is optimized using the loss function 
\begin{align*}
L_{(s, a, s)}(\theta) &= [\mu_\theta(s, a) - s']^T \Sigma_\theta^{-1}(s, a)[\mu_\theta(s, a) - s'] \\
&\quad + \log\det\Sigma_\theta(s, a),
\end{align*}
where $\Sigma_\theta(s, a)$ is assumed to be diagonal.
For input $(s_i, a_i)$, $\sigma^2(s_i, a_i)$ is the trace of $\Sigma_\theta(s_i, a_i)$.  The range of values for the parameter sweep, and the configuration of the rest of the hyperparameters are presented in Table \ref{tab:acrobot_selective_mve_configuration}.

\begin{figure*}[h]
\includegraphics[width=\textwidth]{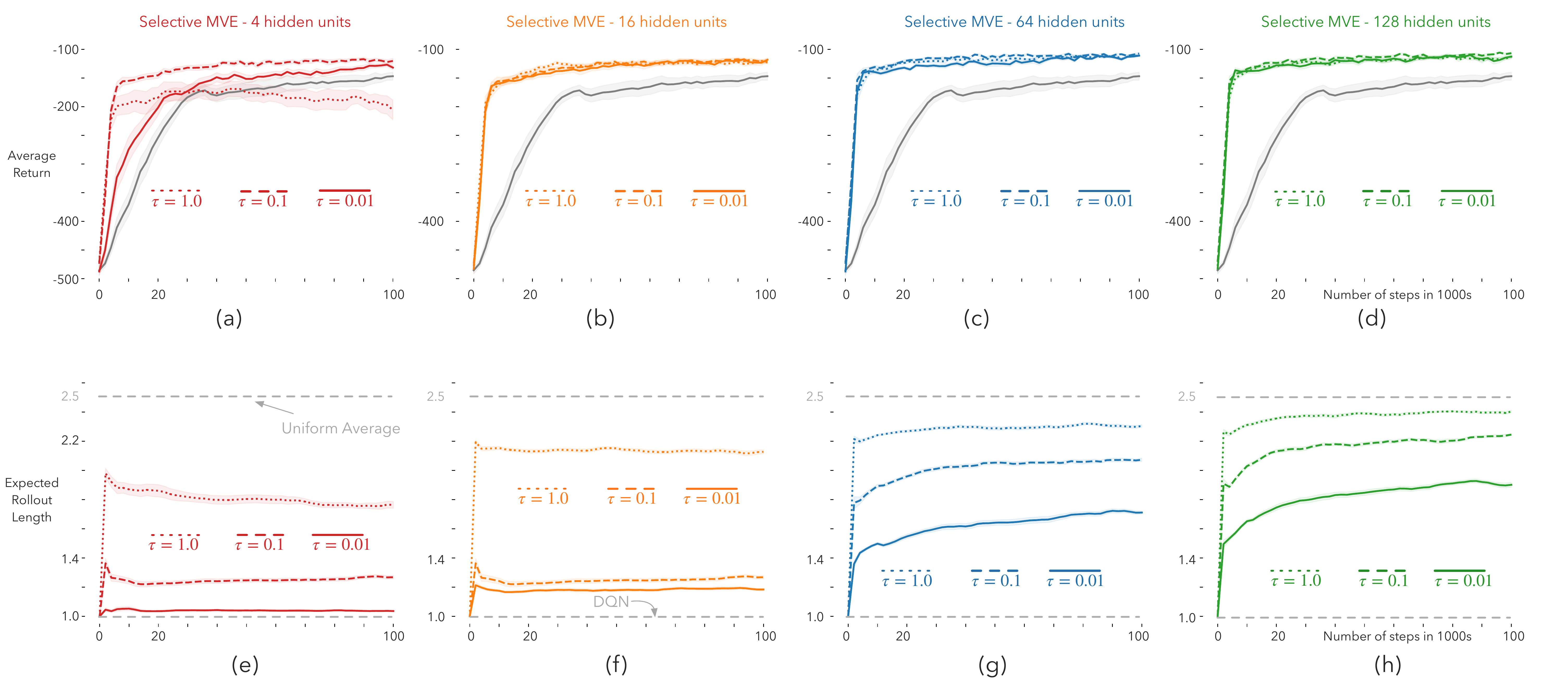}
  \caption{Effect of $\tau$ on the performance of Selective MVE with the value function network consisting of a single hidden layer and 64 hidden units.}
  \label{fig:vf_64h_effect_tau}
\end{figure*}

\begin{figure*}[h]
  \includegraphics[width=\linewidth]{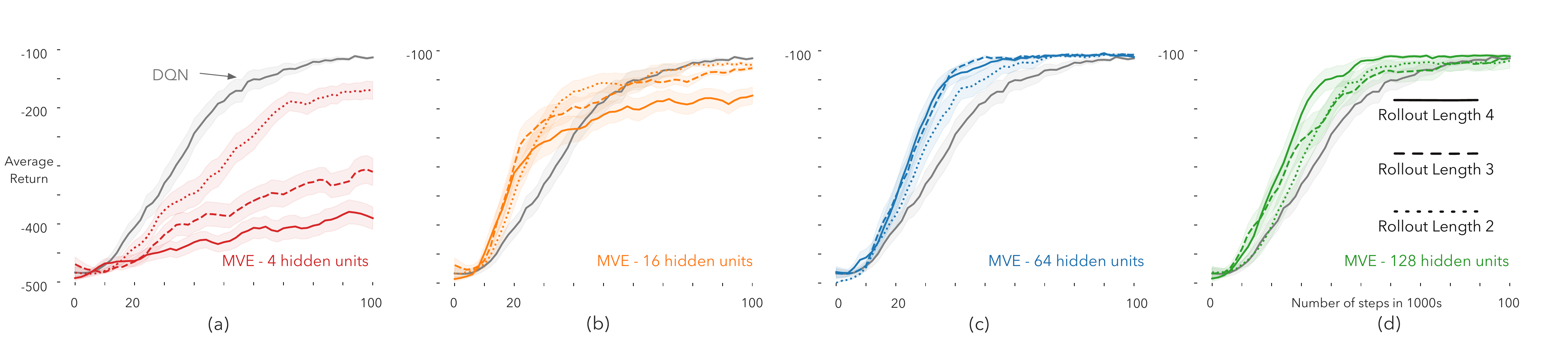}
  \caption{The effect of model capacity on MVE's performance with the value function network consisting of 2 hidden layers with 64 hidden units each.}
  \label{fig:vf_64h_64h_effect_k}
\end{figure*}

\begin{figure*}[h]
  \includegraphics[width=\textwidth]{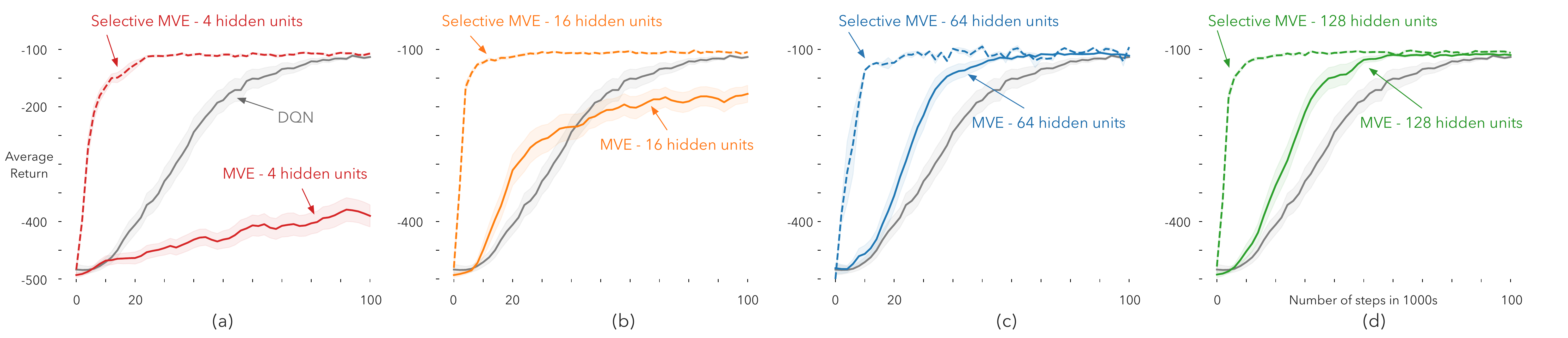}
  \caption{Results of selective MVE ($\tau=0.1$) with value function network of consisting of 2 hidden layers with 64 hidden units each.}
  \label{fig:vf_64h_64h_holistic}
\end{figure*}

\begin{figure*}[h]
  \includegraphics[width=\textwidth]{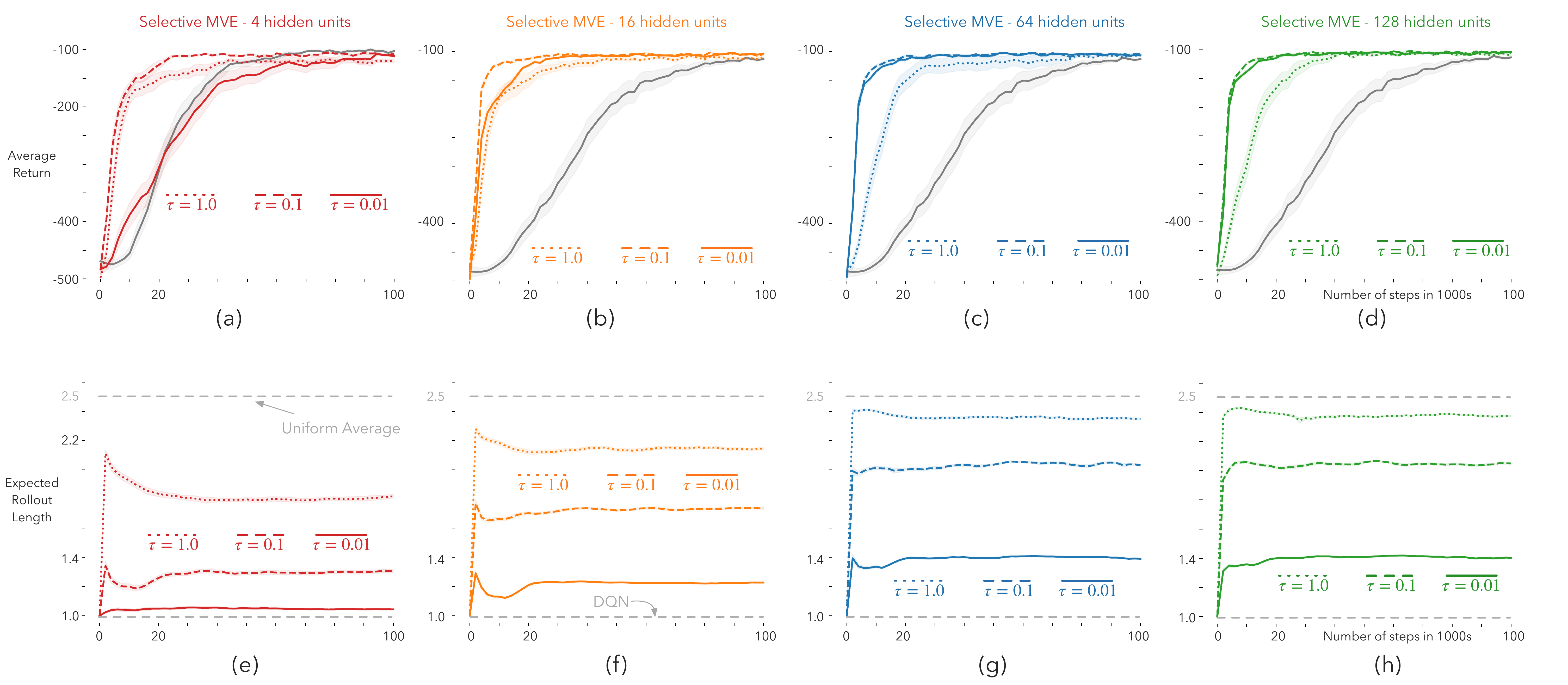}
  \caption{Effect of $\tau$ on the performance of Selective MVE with the value function network consisting of hidden layers with 64 hidden units each.}
  \label{fig:vf_64h_64h_effect_tau}
\end{figure*}

\begin{figure*}[h]
  \includegraphics[width=\linewidth]{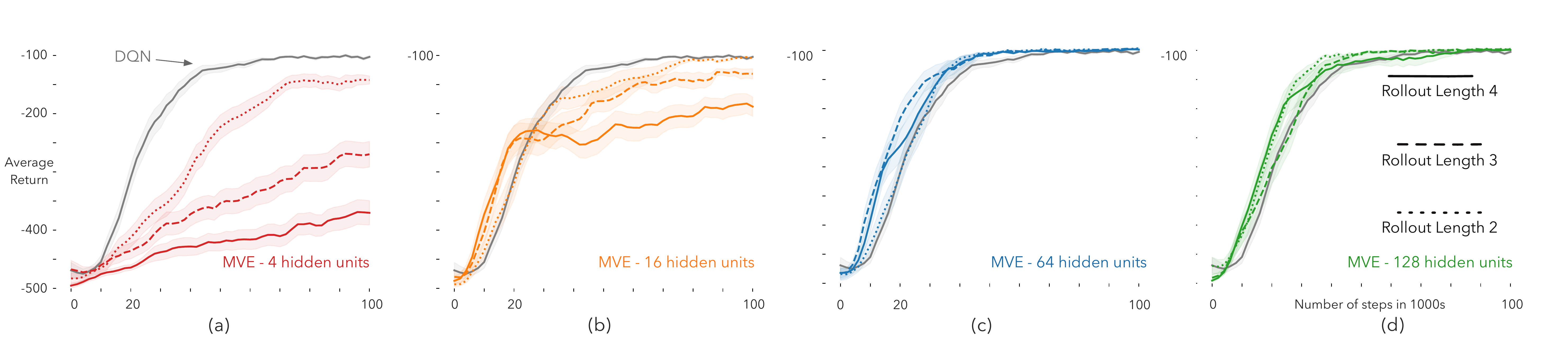}
  \caption{The effect of model capacity on MVE's performance with the value function network consisting of 2 hidden layers with 128 hidden units each.}
  \label{fig:vf_128h_128h_effect_k}
  \end{figure*}

\begin{figure*}[h]
  \includegraphics[width=\textwidth]{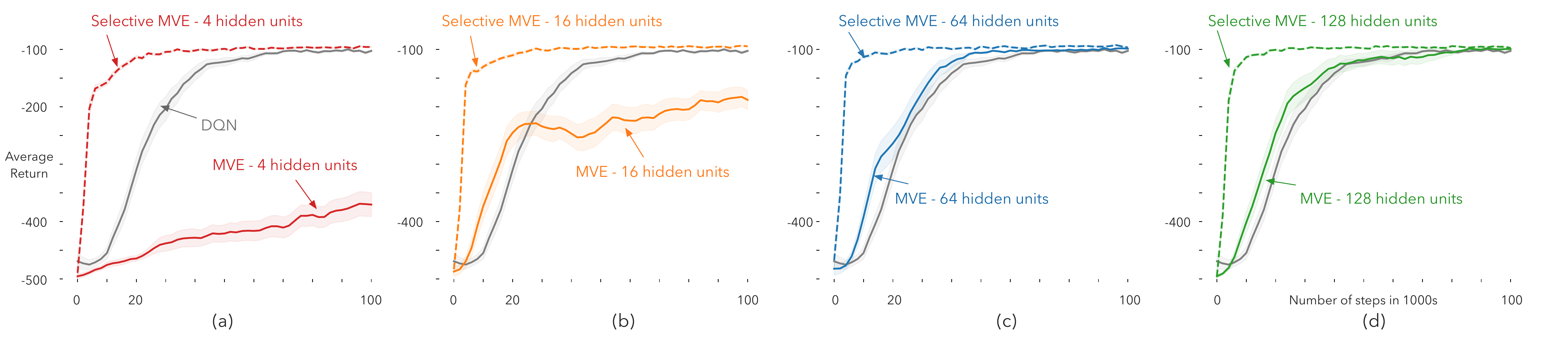}
  \caption{Results of selective MVE ($\tau=0.1$) with the value function network consisting 2 hidden layers with 128 hidden units each.}
  \label{fig:vf_128h_128h_holistic}
  \end{figure*}

\begin{figure*}[h]
  \includegraphics[width=\textwidth]{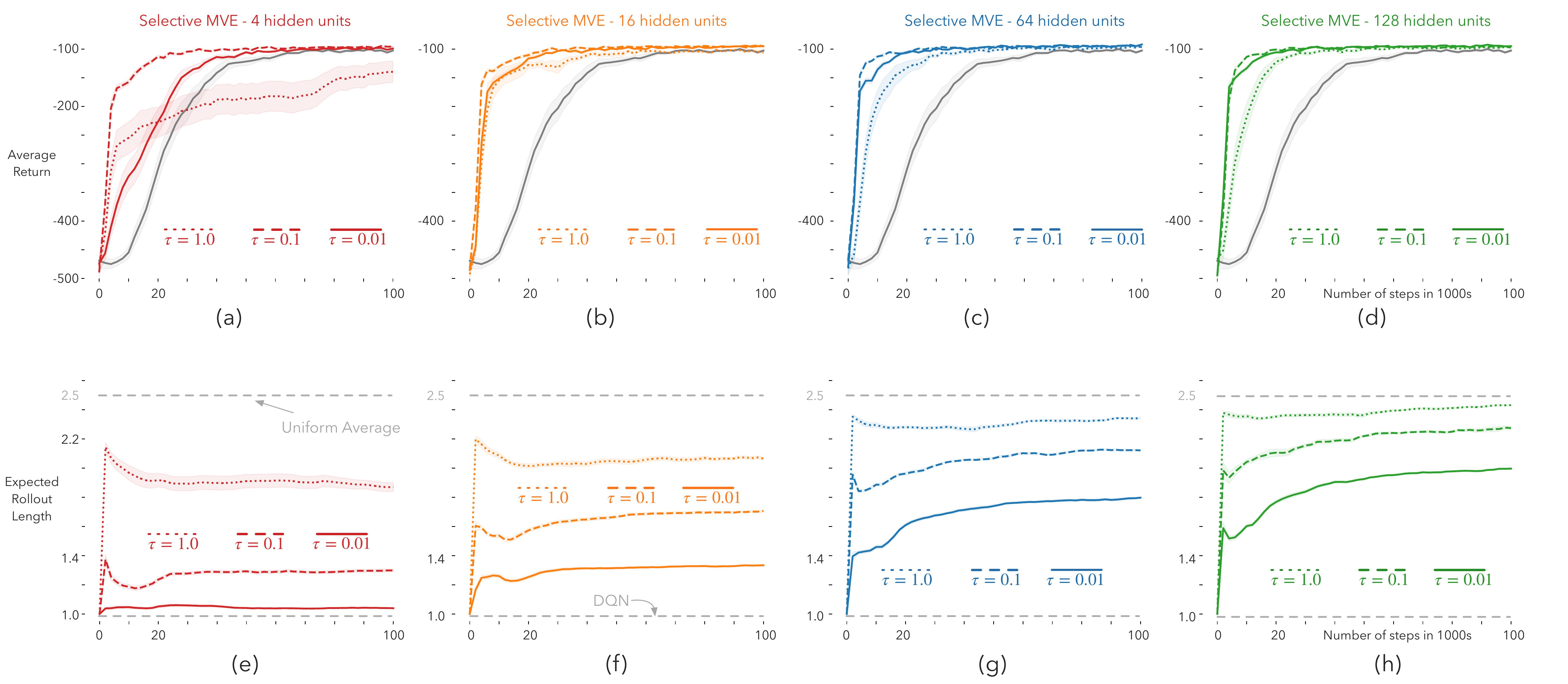}
  \caption{Effect of $\tau$ on the performance of Selective MVE with the value function network consisting of 2 hidden layers with 128 hidden units each.}
  \label{fig:vf_128h_128h_effect_tau}
\end{figure*}

\noindent\textbf{Additional Results for Selective MVE with Learned Variance}.
We present Acrobot results for the value function network architectures with a single hidden layer and 64 hidden units (Figure \ref{fig:vf_64h_effect_k}-\ref{fig:vf_64h_effect_tau}), with 2 hidden layers and 64 hidden units each (Figure \ref{fig:vf_64h_64h_effect_k}-\ref{fig:vf_64h_64h_effect_tau}), and with a 2 hidden layer with 128 hidden units each (Figure \ref{fig:vf_128h_128h_effect_k}-\ref{fig:vf_128h_128h_effect_tau}). All reported curves are obtained by averaging 30 runs; the shaded regions represent the standard error. 
These results are consistent with the discussion in the main paper and provide additional evidence for the utility of learned variance for selective planning. For instance, they suggest that selective planning is useful even when the value function itself has restricted capacity---the network with only 64 hidden units, for example. 

\noindent\textbf{Selective MVE with ensemble variance}. The ensemble-based selective MVE is exactly like the learned-variance variant except for one difference: the uncertainty, $\sigma(s, a)$, is the variance of the predictions made by the individual members of the ensemble. (We add the components of the variance vector to obtain a single number.) The range of values for the parameter sweep, and the configuration of the rest of the hyperparameters are presented in Table \ref{tab:acrobot_selective_mve_ensemble_configuration}.

\section{Regression Example: Additional Details and Results}\label{app:regression}

In Section 4's regression experiment, we use Adam optimizer \citep{kingma2014adam} for all the methods. 
We set the batch size to 16.
We consider the learning rates 0.01, 0.001, and 0.0001. 
We use ReLU activations for non-linearities, and initialize the networks with Glorot initialization \citep{glorot2010understanding}.

For Monte Carlo dropout, we set the dropout probability $p=0.1$. 
To obtain the variance, we perform 10 stochastic forward passes.

For the ensemble method, we use an ensemble of 10 neural networks.
All networks in the ensemble are trained using the squared-error loss.

For randomized prior functions with bootstrapping, we train each member of the ensemble on a bootstrapped dataset generated from the original dataset by randomly sampling with replacement. 

For heteroscedastic regression, we train separate neural networks for the mean and the variance, and optimize them jointly using the loss from Equation 1 (main paper). 
While we change the capacity of the mean network across the three regimes (large network, medium-sized network, and small network), we restrict the variance network to be small---a single hidden layer with 64 hidden units---in all three regimes.\\

\begin{figure*}[h]
\begin{center}
\includegraphics[width=\linewidth]{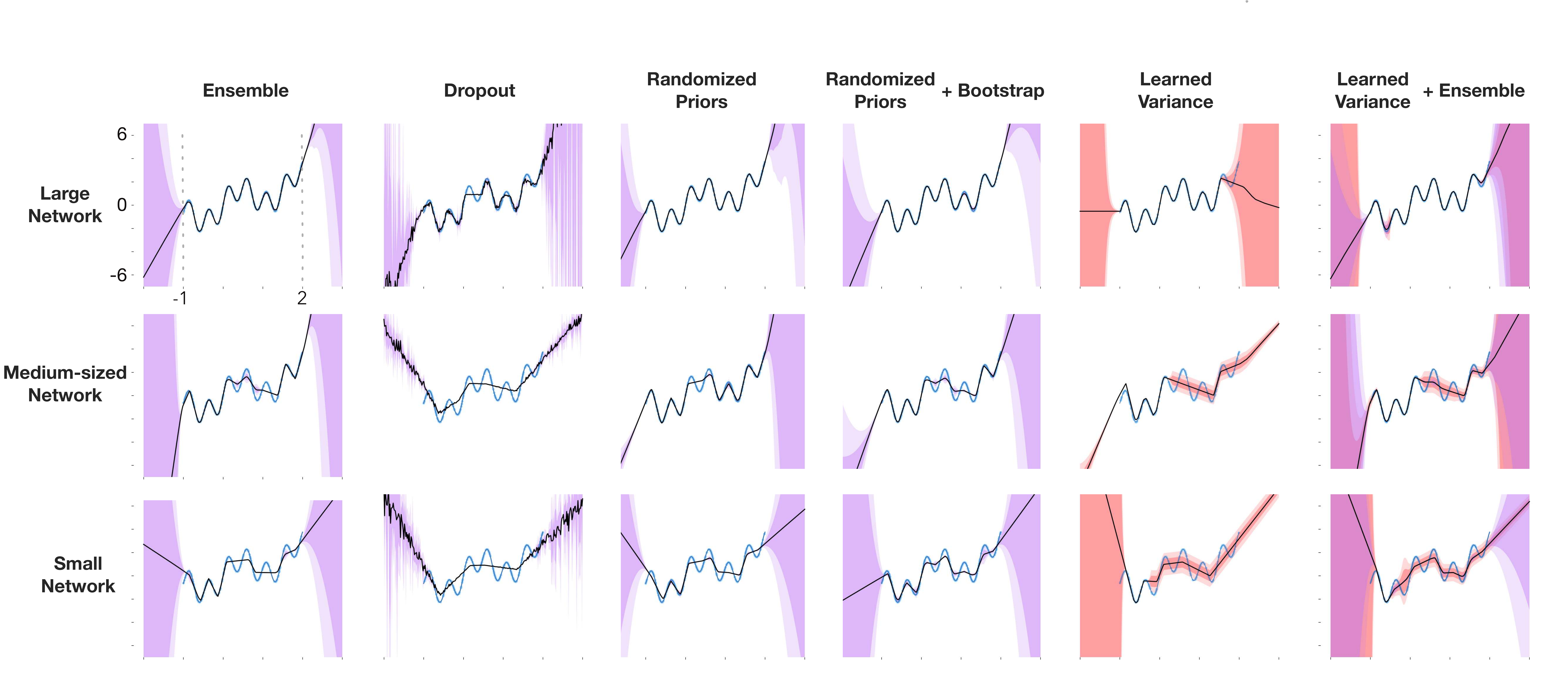}
\caption{Regression results for the learning rate 0.01. 
}\label{fig:regression_results2}
\end{center}
\end{figure*}

\begin{figure*}
\begin{center}
\includegraphics[width=\linewidth]{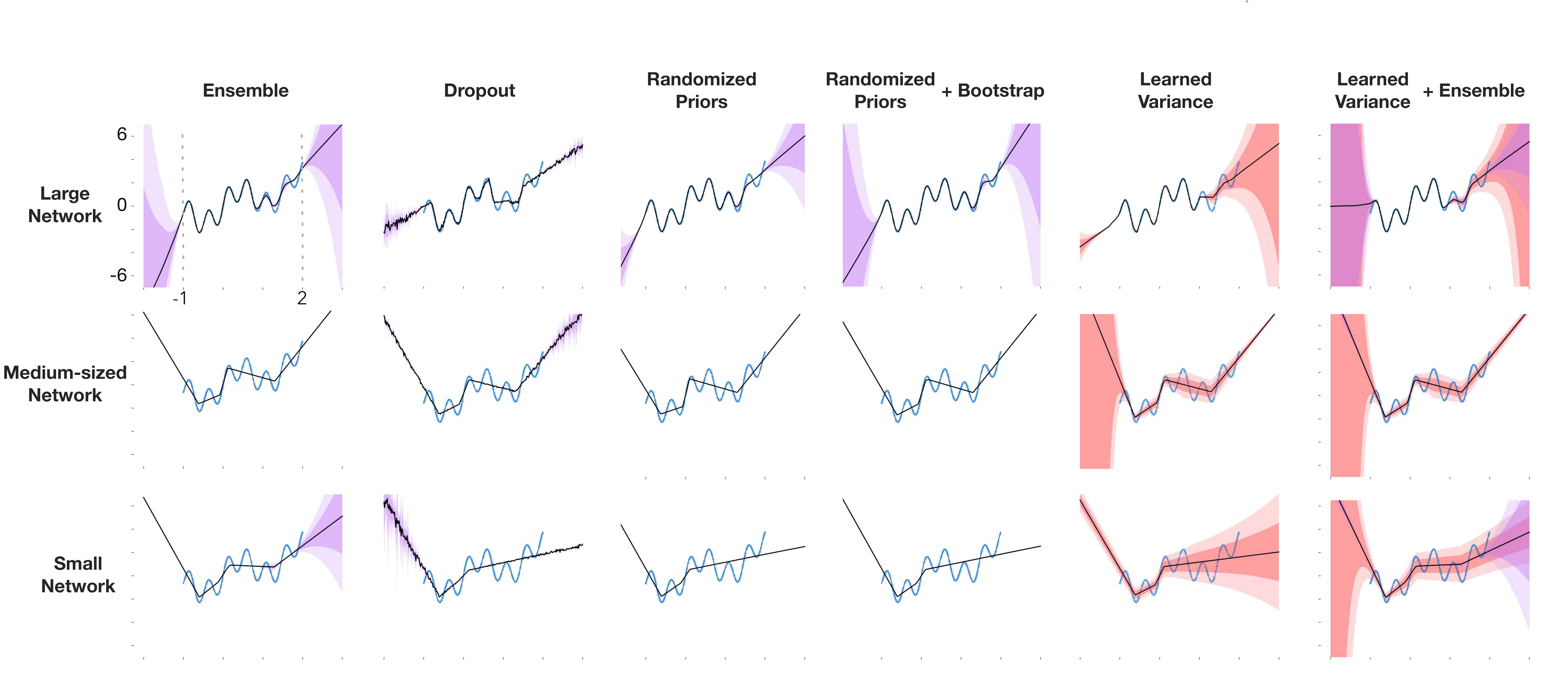}
\caption{Regression results for the learning rate 0.0001.
}\label{fig:regression_results3}
\end{center}
\end{figure*}

For each uncertainty method, every configuration is evaluated using 5 independent runs initialized with a different random seed. While the results remain consistent across the independent runs, we present results for a single run chosen randomly. We present results for additional configurations of the learning rate in Figure \ref{fig:regression_results2}-\ref{fig:regression_results3}.


\newpage


\begin{table*}[h]
\centering
\caption[DQN Hyperparameter Configuration in Acrobot] {DQN hyperparameters used in the experiments. The step-size, the batch size, and the replay memory size were determined by sweeping over the range specified in the respective rows.}
\begin{tabular}{@{}lc@{}}
\toprule
\multicolumn{1}{c}{\textbf{Hyperparameter}} & \multicolumn{1}{c}{\textbf{Values}}      \\\midrule
Optimizer                                   & RMSProp                                  \\
Step-size ($\alpha$)                        & 0.03, 0.01, 0.003, 0.001, 0.0003, 0.0001 \\
Batch size                                  & 16, 32, 64                               \\
Replay memory size                          & 10000, 20000, 50000                      \\
Target network update frequency             & 256 environment steps                    \\
Training frequency                          & 1 update for every environment step      \\
Exploration rate ($\epsilon$)               & 0.1                                      \\
Discount factor ($\gamma$)                  & 1.0                                      \\
\bottomrule
\end{tabular}
\label{tab:acrobot_dqn_configuration}
\end{table*}

\begin{table*}[h]
\centering
\caption[MVE Hyperparameter Configuration in Acrobot] {MVE specific hyperparameters. For each simulated trajectory length (rollout length), the model learning step-size ($\beta$) was determined by sweeping over the range specified in the respective row.}
\begin{tabular}{@{}lc@{}}
\toprule
\multicolumn{1}{c}{\textbf{Hyperparameter}} & \multicolumn{1}{c}{\textbf{Values}} \\ \midrule
Optimizer & Adam \\
Model learning step-size ($\beta$) & 0.1, 0.01, 0.001, 0.0001 \\
Batch size & 16 \\
Loss function & Homoscedastic (MSE) \\
Model learning frequency & 1 update for every environment step  \\
Simulated trajectory length & 2, 3, 4  \\
\bottomrule
\end{tabular}
\label{tab:acrobot_mve_configuration}
\end{table*}

\begin{table*}[h]
\centering
\caption[Selective MVE Hyperparameter Configuration in Acrobot] {Hyperparameters for Selective-MVE. Model learning step-size ($\beta$) was determined by sweeping over the range specified in the respective row.}
\begin{tabular}{@{}lc@{}}
\toprule
\multicolumn{1}{c}{\textbf{Hyperparameter}} & \multicolumn{1}{c}{\textbf{Values}} \\ \midrule
Optimizer & Adam \\
Model learning step-size ($\beta$) & 0.1, 0.01, 0.001, 0.0001 \\
Batch size & 16 \\
Loss function & Heteroscedastic \\
Model learning frequency & 1 update for every environment step  \\
Simulated trajectory length & 4 \\
Softmax temperature ($\tau$) & 0.1 \\
\bottomrule
\end{tabular}
\label{tab:acrobot_selective_mve_configuration}
\end{table*}

\begin{table*}[h]
\centering
\caption[Hyperparameter Configuration for Selective MVE with Ensemble Variance in Acrobot] {Hyperparameters for ensemble-based Selective MVE in Acrobot. Model learning step-size ($\beta$) was determined by sweeping over the range specified in the respective row.}
\label{tab:acrobot_selective_mve_ensemble_configuration}
\begin{tabular}{@{}lc@{}}
\toprule
\multicolumn{1}{c}{\textbf{Hyperparameter}} & \multicolumn{1}{c}{\textbf{Values}} \\ \midrule
Optimizer & Adam \\
Model learning step-size ($\beta$) & 0.1, 0.01, 0.001, 0.0001 \\
Batch size & 16 \\
Loss function & Homoscedastic (MSE) \\
Model learning frequency & 1 update for every environment step  \\
Simulated trajectory length & 4 \\
Softmax temperature ($\tau$) & 0.1 \\
Number of networks & 5 \\
\bottomrule
\end{tabular}
\end{table*}


\end{document}